\documentclass[11pt,a4paper]{article}
\usepackage[hyperref]{naaclhlt2019}
\usepackage{times}
\usepackage{latexsym}

\usepackage{epsfig}
\usepackage{graphicx}
\usepackage{amsmath}
\usepackage{amssymb}
\usepackage{tabularx}
\usepackage{makecell} 
\usepackage{multirow}

\usepackage{color,xcolor}
\usepackage{epsfig}
\usepackage{graphicx}

\usepackage{adjustbox}
\usepackage{array}
\usepackage{booktabs}
\usepackage{colortbl}
\usepackage{float,wrapfig}
\usepackage{hhline}
\usepackage{multirow}
\usepackage{subfigure}

\usepackage{amsmath,amsfonts,amssymb}
\usepackage{bm}
\usepackage{nicefrac}
\usepackage{microtype}

\usepackage{changepage}
\usepackage{extramarks}
\usepackage{fancyhdr}
\usepackage{lastpage}
\usepackage{setspace}
\usepackage{soul}
\usepackage{xspace}

\usepackage{url}

\aclfinalcopy 

\title{UniVSE: Robust Visual Semantic Embeddings via\\ Structured Semantic Representations\footnotemark[1]\thanks{~~This paper is to be presented in the non-archival track of NAACL 2019 Spatial Language Understanding (SpLU) \& Grounded Communication for Robotics (RoboNLP) workshops only. The full version of this paper can be accessed in https://arxiv.org/abs/1904.05521v1, which has been published in the proceedings of CVPR 2019.}}
\author{Hao Wu\textsuperscript{1,3,4,6,\footnotemark[2]\thanks{~~indicates equal contribution.}~~,\thanks{~Work was done when HW, JM and YZ were intern researchers at the Bytedance AI Lab.}}, Jiayuan Mao\textsuperscript{5,6,\footnotemark[2]~~,\footnotemark[3]}, Yufeng Zhang\textsuperscript{2,6,\footnotemark[3]}, Yuning Jiang\textsuperscript{6}, Lei Li\textsuperscript{6}, Weiwei Sun\textsuperscript{1,3,4}, Wei-Ying Ma\textsuperscript{6}\\
	\\
	\textsuperscript{1}School of Computer Science, \textsuperscript{2}School of Economics, Fudan University\\
	\textsuperscript{3}Systems and Shanghai Key Laboratory of Data Science, Fudan University\\
	\textsuperscript{4}Shanghai Insitute of Intelligent Electroics \& Systems\\
	\textsuperscript{5}ITCS, Institute for Interdisciplinary Information Sciences, Tsinghua University, \textsuperscript{6}Bytedance AI Lab\\
	{\tt\small \{wuhao5688, zhangyf, wwsun\}@fudan.edu.cn,  mjy14@mails.tsinghua.edu.cn,}\\
	{\tt\small \{jiangyuning, lileilab, maweiying\}@bytedance.com}
}
\date{}

\definecolor{MyDarkBlue}{rgb}{0,0.08,1}
\definecolor{MyDarkGreen}{rgb}{0.02,0.6,0.02}
\definecolor{MyDarkRed}{rgb}{0.8,0.02,0.02}
\definecolor{MyDarkOrange}{rgb}{0.40,0.2,0.02}
\definecolor{MyPurple}{RGB}{111,0,255}
\definecolor{MyRed}{rgb}{1.0,0.0,0.0}
\definecolor{MyGold}{rgb}{0.75,0.6,0.12}
\definecolor{MyDarkgray}{rgb}{0.66, 0.66, 0.66}

\newcommand{\model}{Unified VSE\xspace} 
\newcommand{\modelshort}{UniVSE\xspace}
\newcommand{\modelshortnopostspace}{UniVSE}

\newcommand{\imgglobal}{\mathbf{v}}
\newcommand{\imglocal}{\mathbf{V}}
\newcommand{\txtglobal}{\mathbf{u}}
\newcommand{\txtobj}{\mathbf{u}_o}

\newcommand{\wordvec}{\mathbf{w}}

\newcommand{\margin}{\delta}

\newcommand{\Eg}{E.g.\xspace}
\newcommand{\eg}{e.g.\xspace}
\newcommand{\ie}{i.e.\xspace}
\newcommand{\wrt}{w.r.t.\xspace}
\newcommand{\subsectionnoid}[1]{\noindent\textbf{#1}}

\begin{document}
\maketitle
\begin{abstract}  
	We propose Unified Visual-Semantic Embeddings (UniVSE) for learning a joint space of visual and textual concepts. The space unifies the concepts at different levels, including objects, attributes, relations, and full scenes. A contrastive learning approach is proposed for the fine-grained alignment from only image-caption pairs. Moreover, we present an effective approach for enforcing the coverage of semantic components that appear in the sentence. We demonstrate the robustness of \model in defending text-domain adversarial attacks on cross-modal retrieval tasks. Such robustness also empowers the use of visual cues to resolve word dependencies in novel sentences.
\end{abstract}

\vspace{-1em}

\section{Introduction}
\begin{figure}[t]
	\centering
	\includegraphics[width=80mm]{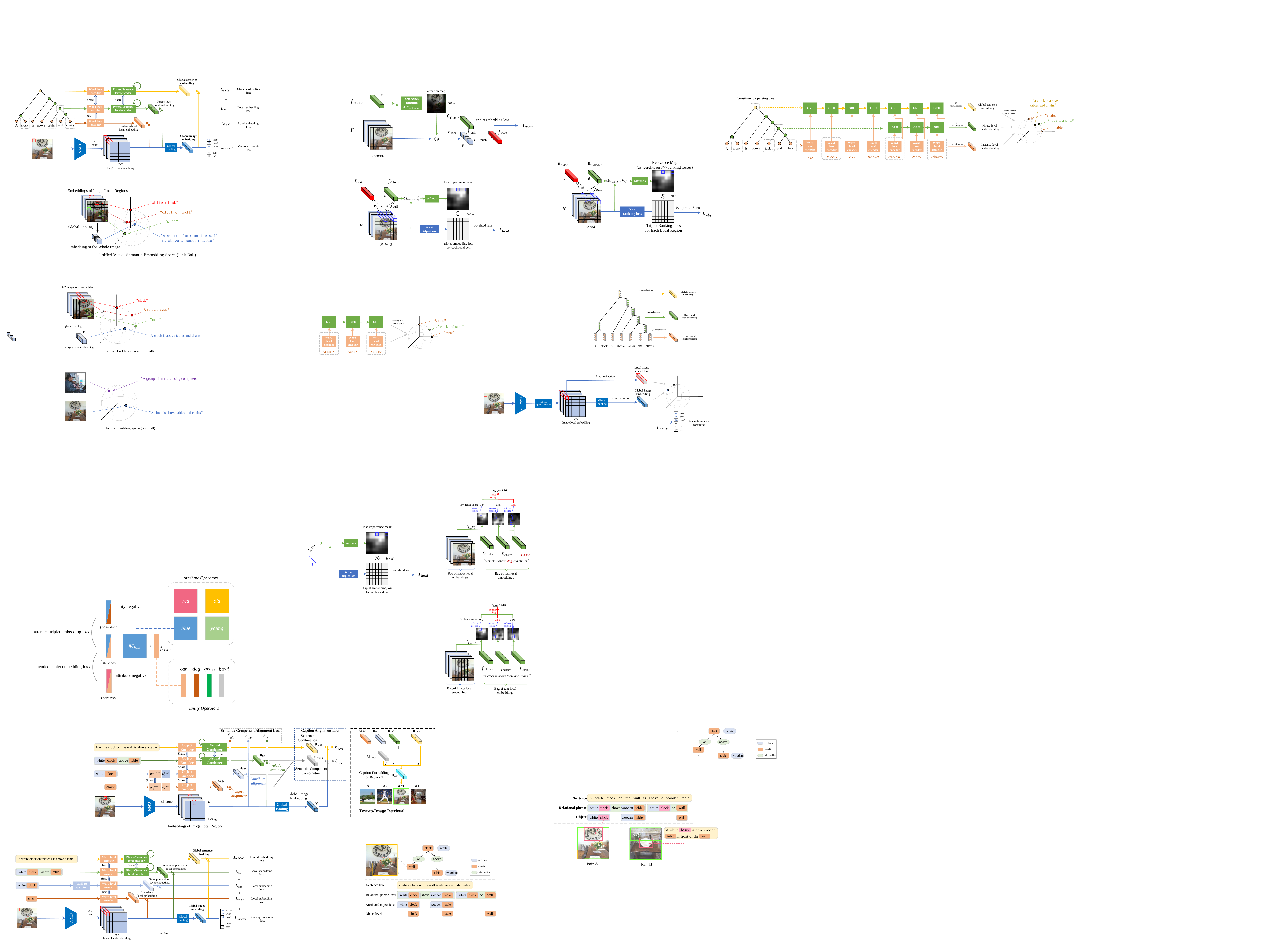}
	\caption{Two examplar image-caption pairs. Humans are able to establish accurate and generalizable alignments between vision and language, at different levels: objects, relations and full sentences. Pair A and B form a pair of contrastive example for the concepts \texttt{clock} and \texttt{basin}.}
	\label{fig:sng}
	\vspace{-0.1in}
\end{figure}

    We study the problem of establishing accurate and generalizable alignments between visual concepts and textual semantics efficiently, based upon rich but few, paired but noisy, or even biased visual-textual inputs (\eg, image-caption pairs). 
    Consider the image-caption pair A shown in Fig.~\ref{fig:sng}: {\it ``A white clock on the wall is above a wooden table''}.
    The alignments are formed at multiple levels: 
    This short sentence can be decomposed into a rich set of semantic components \cite{bootstrapping}: objects ({\tt clock}, {\tt table} and {\tt wall}) and relations (clock {\tt above} table, and clock {\tt on} wall). These components are linked with different parts of the scene.

\begin{figure}
		\centering
		\includegraphics[width=80mm]{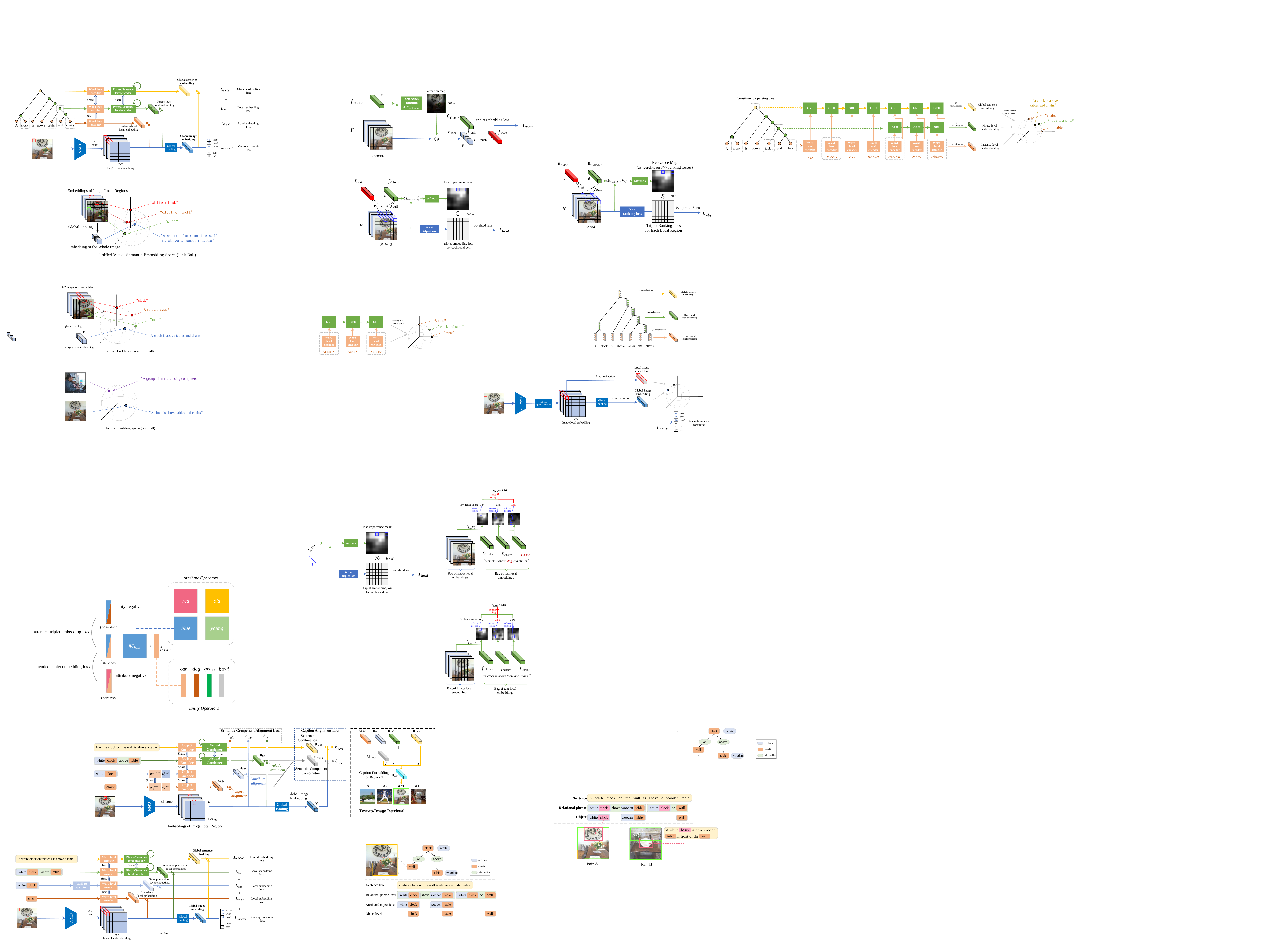}
		\caption{We build a visual-semantic embedding space, which unifies the embeddings for objects, attributes, relations and full scenes.}
		\label{fig:unit_ball}
		\vspace{-0.1in}
\end{figure}

    This motives our work to introduce {\it Unified Visual-Semantic Embeddings} ({\it Unified VSE} for short) Shown in Fig.~\ref{fig:unit_ball}, \model bridges visual and textual representation in a joint embedding space that unifies the embeddings for objects (noun phrases vs. visual objects), attributes (prenominal phrases vs. visual attributes), relations (verbs or prepositional phrases vs. visual relations) and scenes (sentence vs. image).

    There are two major challenges in establishing such a factorized alignment. First, the link between the textual description of an object and the corresponding image region is ambiguous: A visual scene consists of multiple objects, and thus it is unclear to the learner which object should be aligned with the description.
    Second, it could be problematic to directly learn a neural network that combines various semantic components in a caption and form an encoding for the full sentence, with the training objective to maximize the cross-modal retrieval performance in the training set (\eg, in \cite{cse,mcnn,vsec}).
    As reported by \cite{vsec}, because of the inevitable bias in the dataset (\eg, two objects may co-occur with each other in most cases, see the table and the wall in Fig. \ref{fig:sng} as an example), the learned sentence encoders usually pay attention to only part of the sentence. As a result, they are vulnerable to text-domain adversarial attacks: Adversarial captions constructed from original captions by adding small perturbations (\eg, by changing {\tt wall} to be {\tt shelf}) can easily fool the model \cite{vsec, foilit}.

    We resolve the aforementioned challenges by a natural combination of two ideas: {\it cross-situational learning} and the enforcement of {\it semantic coverage} that regularizes the encoder.
    Cross-situational learning, or learning from contrastive examples \cite{crosssituational}, uses contrastive examples in the dataset to resolve the referential ambiguity of objects: Looking at both Pair A and B in Fig. \ref{fig:sng}, we know that {\tt Clock} should refer to an object that occurs only in scene A but not B.
    Meanwhile, to alleviate the biases of datasets such as object co-occurrence, we present an effective approach that enforces the {\it semantic converage}: The meaning of a caption is a composition of all semantic components in the sentence \cite{bootstrapping}. Reflectively, the embedding of a caption should have a coverage of all semantic components, while changing any of them should affect the global caption embedding.

    Conceptually and empirically, \model makes the following three contributions.
    
    First, the explicit factorization of the visual-semantic embedding space enables us to build a fine-grained correspondence between visual and textual data, which further benefits a set of downstream visual-textual tasks. We achieve this through a contrastive example mining technique that uniformly applies to different semantic components, in contrast to the sentence or image-level contrastive samples used by existing visual-semantic learning \cite{cse,mcnn,vsepp}.

    Second, we propose a caption encoder that ensures a coverage of all semantic components appeared in the sentence. We show that this regularization helps our model to learn a robust semantic representation for captions. It effectively defends adversarial attacks on the text domain.

    Furthermore, we show how our learned embeddings can provide visual cues to assist the parsing of novel sentences, including determining content word dependencies and labelling semantic roles for certain verbs. It ends up that our model can build reliable connections between vision and language using given semantic cues and in return, bootstrap the acquisition of language.

\section{Related work}

\subsectionnoid{Visual semantic embedding.}
Visual semantic embedding \cite{vse} is a common technique for learning a joint representation of vision and language. The embedding space empowers a set of cross-modal tasks such as image captioning \cite{vinyals2015show,xu2015show,donahue2015long} and visual question answering \cite{vqa,xu2016ask}. 

A fundamental technique proposed in \cite{vse} for aligning two modalities is to use the pairwise ranking to learn a distance metric from similar and dissimilar cross-modal pairs \cite{deepsp, gvse, scnlm, 2waynet, rrfnet, mnlm}.
As a representative, VSE++ \cite{vsepp} uses the online hard negative mining (OHEM) strategy \cite{ohem} for data sampling and shows the performance gain.
VSE-C \cite{vsec}, based on VSE++, enhances the robustness of the learned visual-semantic embeddings by incorporating rule-generated textual adversarial samples as hard negatives during training. In this paper, we present a contrastive learning approach based on semantic components. 

There are multiple VSE approaches that also use linguistically-aware techniques for the sentence encoding and learning.
Hierarchical multimodal LSTM (HM-LSTM) \cite{hmlstm} and \cite{attconstraint}, as two examples, both leverage the constituency parsing tree. 
Multimodal-CNN (m-CNN) \cite{mcnn} and CSE \cite{cse} apply convolutional neural networks to the caption and extract the a hierarchical representation of sentences.
Our model differs with them in two aspects. First, \model is built upon a factorized semantic space instead of the syntactic knowledge. Second, we employ a contrastive example mining approach that uniformly applies to different semantic components. It substantially improves the learned embeddings, while the related works use only sentence-level contrastive examples.

The learning of object-level alignment in unified VSE is also related to \cite{deepvs, defrag, mivse}, where the authors incorporate pre-trained object detectors for the semantic alignment. \cite{beans} propose a selective pooling technique for the aggregation of object features. Compared with them, \model presents a more general approach that embeds concepts of different levels, while still requiring no extra supervisions.

\subsectionnoid{Structured representation for vision and language.}
We connect visual and textual representations in a structured embedding space. The design of its structure is partially motivated by the papers on relational visual representations (scene graphs) \cite{vrd,sng_retrieval, sng_gen}, where a scene is represented by a set of objects and their relations. Compared with them, our model does not rely on labelled graphs during training.

Researchers have designed various types of representations \cite{amr, ug} as well as different models \cite{dependencygramma, ccg} for translating natural language sentences into structured representations. In this paper, we present how the usage of such semantic parsing into visual-semantic embedding facilitates the learning of the embedding space. Moreover, we present how the learned VSE can, in return, helps the parser to resolve parsing ambiguities using visual cues.

\section{Unified Visual-Semantic Embeddings}

We now describe the overall architecture and training paradigm for the proposed {\it Unified Visual-Semantic Embeddings}.
Given an image-caption pair, we first parse the caption into a structured meaning representation, composed by a set of semantic components: \emph{object nouns}, \emph{prenominal modifiers}, and \emph{relational dependencies}. We encode different types of semantic components with type-specific encoders. A caption encoder combines the embedding of the semantic components into a caption semantic embedding. Jointly, we encode images with a convolutional neural network (CNN) into the same, {\it unified VSE} space. The distance between the image embedding and the sentential embedding measures the semantic similarity between the image and the caption.

We employ a multi-task learning approach for the joint learning of embeddings for semantic components (as the ``basis'' of the VSE space) as well as the caption encoder (as the combiner of semantic components).

\subsection{Visual-Semantic Embedding: A Revisit}\label{sec:global_learning}
We begin the section with an introduction to the two-stream VSE approach. It jointly learns the embedding spaces of two modalities: vision and language, and aligns them using parallel image-text pairs (\eg, image and captions from the MS-COCO dataset \cite{mscoco}).

Let $\imgglobal \in \mathbb{R}^d$ be the representation of the image
and $\txtglobal \in \mathbb{R}^d$ be the representation of a caption matching this image, both encoded by neural modules.
To archive the alignment, a bidirectional margin-based ranking loss has been widely applied \cite{vsepp, cse, smlstm}.
Formally, for an image (caption) embedding $\imgglobal$ ($\txtglobal$), denote the embedding of its matched caption (image) as $\txtglobal^+$ ($\imgglobal^+$). A negative (unmatched) caption (image) is sampled whose embedding is denoted as $\txtglobal^-$ ($\imgglobal^-$). We define the bidirectional ranking loss $\ell_{sent}$ between captions and images as:

\vspace{-0.5em}
{\small
\begin{align}\notag
\ell_{sent} = & \sum_{\txtglobal} F_{\imgglobal^-}\left(|\margin + s(\txtglobal,\imgglobal^-) - s(\txtglobal,\imgglobal^+) |_+\right) \\
+ & \sum_{\imgglobal} F_{\txtglobal^-}\left(|\margin + s(\txtglobal^-,\imgglobal) - s(\txtglobal^+,\imgglobal) |_+\right)
\label{eq:vse}
\vspace{-1.5em}
\end{align}}
where $\margin$ is a predefined margin, $|x|_+ = \max(x, 0)$ is the traditional ranking loss and $F_\mathbf{x}(\cdot) = \max_{\mathbf{x}}(\cdot)$ denotes the hard negative mining strategy \cite{vsepp, ohem}. $s(\cdot,\cdot)$ is a similarity function between two embeddings and is usually implemented as cosine similarity \cite{vsepp, vsec, cse}.

\subsection{Semantic Encodings}\label{sec:local_learning}

The encoding of a caption is made up of three steps. As an example, consider the caption
 {\it``A white clock on the wall is above a wooden table''}. 1) We extract a structured meaning representation as a collection of three types of semantic components: object (\verb`clock`, \verb`wall`, \verb`table`), attribute-object dependencies (\verb`white` clock, \verb`wooden` table) and relational dependencies (clock \verb`above` table, clock \verb`on` wall). 2) We encode each component as well as the full sentence with type-specific encoders into the unified VSE space. 3)
We represent the embedding of the caption by combining the semantic components.


\subsectionnoid{Semantic parsing.}
We implement a semantic parser \footnote{https://github.com/vacancy/SceneGraphParser} of image captions based on \cite{snggen}. Given the input sentence, the parser first performs a syntactic dependency parsing. A set of rules is applied to the dependency tree and extracts object entities appeared in the sentence, adjectives that modify the object nouns, subjects/objects of the verbs and prepositional phrases.
For simplicity, we consider only single-word nouns for objects and single-word adjectives for object attributes.

\subsectionnoid{Encoding objects and attributes.}
We use an unified object encoder $\phi$ for nouns and adjective-noun pairs. For each word $w$ in the vocabulary, we initialize a basic semantic embedding $\wordvec^{(basic)}\in\mathbb{R}^{d_{basic}}$ and a modifier semantic embedding $\wordvec^{(modif)} \in \mathbb{R}^{d_{modif}}$.

For a single noun word $w_n$ (\eg, \verb`clock`), we define its embedding $\mathbf{w}_n$ as $\wordvec_n^{(basic)} \oplus \wordvec_n^{(modif)}$, where $\oplus$ means the concatenation of vectors. For an (adjective, noun) pair $(w_a, w_n)$ (\eg, (\verb`white`, \verb`clock`)), its embedding $\mathbf{w}_{a,n}$ is defined as $\wordvec_n^{(basic)} \oplus \wordvec_a^{(modif)}$ where $ \wordvec_a^{(modif)}$ encodes the attribute information.
In implementation, the basic semantic embedding is initialized from GloVe \cite{glove}. The modifier semantic embeddings (both $\wordvec_n^{(modif)}$ and $\wordvec_a^{(modif)}$) are randomly initialized and jointly learned. $\wordvec_n^{(modif)}$ can be regarded as an intrinsic modifier for each nouns.

To fuse the embeddings of basic and modifier semantics, we employ a gated fusion function:
{\footnotesize
\begin{align*}
\phi(\wordvec_n) &= \mathrm{Norm}(\sigma(\mathbf{W}_1\wordvec_n+\mathbf{b}_1))\tanh(\mathbf{W}_2\wordvec_n+\mathbf{b}_2)),\\
\phi(\wordvec_{a, n}) &=\mathrm{Norm}(\sigma(\mathbf{W}_1\wordvec_{a, n}+\mathbf{b}_1)\tanh(\mathbf{W}_2\wordvec_{a, n}+\mathbf{b}_2)).
\end{align*}
}
Throughout the text, $\sigma$ denotes the sigmoid function: $\sigma(x) = 1/(1 + \exp (-x))$, and $\mathrm{Norm}$ denotes the L2 normalization, \ie, $\mathrm{Norm}(\wordvec) = \wordvec / \|\wordvec\|_2 $. One may interpret $\phi$ as a GRU cell \cite{gru} taking no historical state.

\subsectionnoid{Encoding relations and full sentence.}
Since relations and sentences are the composed based on objects, we encode them with a neural combiner $\psi$, which takes the embeddings of word-level semantics encoded by $\phi$ as input. In practice, we implement $\psi$ as an uni-directional GRU \cite{gru}, and pick the L2-normalized last state as the output.

To obtain a visual-semantic embedding for a relational triple $(w_s, w_r, w_o)$ (\eg, (\verb`clock`, \verb`above`, \verb`table`)), we first extract the word embeddings for the subject, relational word and the object using $\phi$. We then feed the encoded word embeddings in the same order into $\psi$ and takes the L2-normalized last state of the GRU cell. Mathematically, $\txtglobal_{rel} = \psi(w_s, w_r, w_o) = \psi(\{ \phi(\wordvec_s), \phi(\wordvec_r), \phi(\wordvec_o) \})$.

The embedding of a sentence $\txtglobal_{sent}$ is computed over the word sequence $w_1, w_2, \cdots w_k$ of the caption:
\\[-1.5em]\begin{center} $\txtglobal_{sent} = \psi(\{ \phi(\mathbf{w}_1), \phi(\mathbf{w}_2), \cdots, \phi(\mathbf{w}_k) \})$,\end{center}
where for any word $x$, $\phi(\wordvec_x) = \phi(\wordvec_x^{(basic)} \oplus \wordvec_x^{(modif)})$
Note that we share the weights of the encoders $\psi$ and $\phi$ among the encoding processes of all semantic levels. This allows our encoders of various types of components to bootstrap the learning of each other.

\subsectionnoid{Combining all of the components.}
A straight-forward implementation of the caption encoder is to directly use the sentence embedding $\txtglobal_{sent}$, as it has already combined the semantics of components in a contextually-weighted manner \cite{levy2018lstm}. However, it has been revealed in \cite{vsec} that such combination is vulnerable to adversarial attacks: Because of the biases in the dataset, the combiner $\psi$ usually focuses on only a small set of semantic components appeared in the caption.

We alleviate such biases by enforcing the coverage of the semantic components appeared in the sentence. Specifically, to form the caption embedding $\txtglobal_{cap}$, the sentence embedding $\txtglobal_{sent}$ is combined with an explicit bag-of-components embedding $\txtglobal_{comp}$. 
Mathematically, we define $\txtglobal_{comp}$ as an unweighted aggregation of all components in the sentence:

{\footnotesize
	\begin{align}\notag
		\txtglobal_{comp} = \mathrm{Norm}\left( \sum\nolimits_{obj} \txtglobal_{obj} + \sum\nolimits_{attr} \txtglobal_{attr} + \sum\nolimits_{rel} \txtglobal_{rel}\right)
	\end{align}}
and encode the caption as: $\txtglobal_{cap} = \alpha \txtglobal_{sent} + (1 - \alpha) \txtglobal_{comp}$, where $0 \le \alpha \le 1$ is a scalar weight. The presence of $\txtglobal_{comp}$ disallows the ignorance of any of the components in the final caption embedding $\txtglobal_{cap}$.


\subsection{Image Encodings}
We use CNN to encode the input RGB image into the unified VSE space. Specifically, we choose a ResNet-152 model \cite{resnet} pretrained on ImageNet \cite{imagenet} as the image encoder. We apply a layer of $1\times 1$ convolution on top of the last convolutaion layer (\ie, \verb'conv5_3') and obtain a convolutional feature map of shape $7 \times 7 \times d$ for each image. $d$ denotes the dimension of the unified VSE space.

The feature map, denoted as $\imglocal \in \mathbb{R}^{7\times7\times d}$, can be view as the embeddings of $7\times 7$ local regions in the image. 
The embedding $\imgglobal$ for the whole image is defined as the aggregation of the embeddings at all regions through a global spatial pooling operator.

\subsection{Learning Paradigm}\label{sec:training_paradigm}
In this section, we present how to align vision and language into the unified space using contrastive learning on different semantic levels. 
We start from the generation of contrastive exampls for different semantic components.

\subsectionnoid{Negative example sampling.}
It has been discussed in \cite{vsec} that to explore a large compositional space of semantics, directly sampling negative captions from a human-built dataset (\eg, MS-COCO captions) is not sufficient. In this paper, instead of manually define rules that augment the training data as in \cite{vsec}, we address this problem by sampling contrastive negative examples in the explicitly factorized semantic space. The generation does not require manually labelled data, and can be easily applied to any datasets. For a specific caption, we generate the following four types of contrastive negative samples.
\begin{itemize}
    \vspace{-0.1in}
    \item {\bf Nouns.} We sample negative noun words from all nouns that do not appear in the caption. \footnote{For the MS-COCO dataset, in all 5 captions associated with the same image. This also applies to other components.}
    \vspace{-0.1in}
    \item {\bf Attribute-noun pairs.} We sample negative pairs by randomly substituting the adjective by another adjective or substituting the noun.
    \vspace{-0.1in}
    \item {\bf Relational triples.} We sample negative triples by randomly substituting the subject, or the relation, or the object. Moreover, we also sample the whole relational triples of captions in the dataset which describe other images, as the negative triples.
    \vspace{-0.1in}
    \item {\bf Sentences.} We sample negative sentences from the whole dataset. Meanwhile, following \cite{vse,vsepp}, we also sample negative images from the whole dataset as contrastive images.
    \vspace{-0.1in}
\end{itemize}

The key motivation behind our visual-semantic alignment is that: an object appears in a local region of the image, while the aggregation of all local regions should be aligned with the full semantics of a caption.

\begin{figure}
	\centering
	\includegraphics[width=75mm]{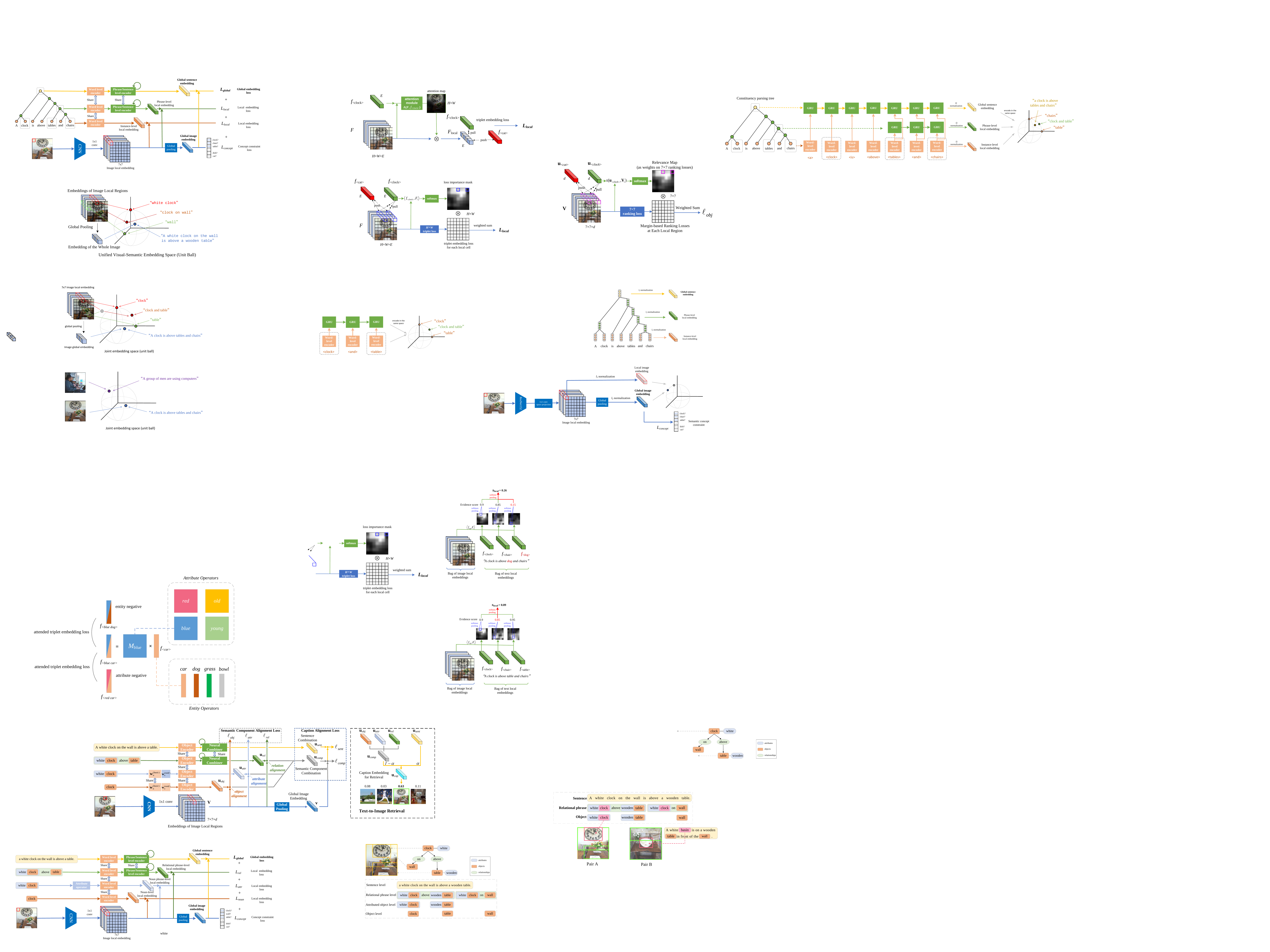}
	\caption{An illustration of our relevance-weighted alignment mechanism. The relevance map shows the similarity of each region with the object embedding $\txtglobal_{<clock>}$. We weight the alignment loss with the map to reinforce the correspondence between the $\txtglobal_{<clock>}$ and its matched region.}
	\label{fig:attention_loss}
	\vspace{-0.15in}
\end{figure}

\subsectionnoid{Local region-level alignment.}
In detail, we propose a relevance-weighted alignment mechanism for linking textual object descriptors and local image regions.
As shown in Fig.~\ref{fig:attention_loss}, consider the embedding of a positive textual object descriptor $\txtobj^{+}$, a negative textual object descriptor $\txtobj^{-}$ and the set image local region embeddings $\imglocal_i$ where $i \in 7\times7$ extracted from the image. 
We generate a relevance map $\mathbf{M} \in \mathbb{R} ^ {7\times7}$ with $\mathbf{M}_{i}, i\in7\times7$ representing the relevance between $\txtobj^+$ and $\imglocal_i$, computed as as Eq.~(\ref{eq:relevance_map}).
We compute the loss for noun and (adjective, noun) pairs by:

\vspace{-0.08in}
{\footnotesize
\begin{align}\label{eq:relevance_map}
\mathbf{M}_i = & \frac{\exp(s(\txtobj^+, \imglocal_i))}{\sum_{j}\exp(s(\txtobj^+,\imglocal_j))}\\
\ell_{obj} =& \sum_{i \in 7 \times 7}\left(\mathbf{M}_i \cdot \left|\margin + s(\txtobj^-, \imglocal_i) - s(\txtobj^+, \imglocal_i)\right|_+\right)
\end{align}}
The intuition behind the definition is that, we explicitly try to align the embedding at each image region with $\txtobj^+$. The losses are weighted by the matching score, thus reinforce the correspondence between $\txtobj^+$ and the matched region. This technique is related to multi-instance learning \cite{dmil}.

\begin{table*}
	\textls[0]{
	\scriptsize
	\begin{center}
		\begin{tabular}{c|ccc|c||ccc|c||ccc|c||c}
			\hline
			& \multicolumn{4}{c||}{\textbf{Object attack}} & \multicolumn{4}{c||}{\textbf{Attribute attack}} & \multicolumn{4}{c||}{\textbf{Relation attack}}\\
			\hline
			Metric & R@1 & R@5 & R@10 & rsum & R@1 & R@5 & R@10 & rsum & R@1 & R@5 & R@10 & rsum & total sum\\			
			\hline				
			VSE++ & 32.3 & 69.6	& 81.4 & 183.3 		&19.8&59.4&76.0&155.2		&26.1&66.8&78.7&171.6		&510.1\\
			VSE-C & 41.1 & 76.0 & 85.6  & 202.7  	&26.7&61.0&74.3&162.0			&35.5&71.1&81.5&188.1		&552.8\\
			\modelshort($\txtglobal_{sent}$+$\txtglobal_{comp}$)&\textbf{45.3}&\textbf{78.3}&\textbf{87.3}&\textbf{210.9}		&\textbf{35.3}&\textbf{71.5}&\textbf{83.1}&\textbf{189.9}	&\textbf{39.0}&\textbf{76.5}&\textbf{86.7}&\textbf{202.2}	&\textbf{603.0}\\
			\hline	
			\modelshort($\txtglobal_{sent}$) & 40.7 &	76.4 & 85.5 & 202.6 &30.0&70.5&80.6&181.1		&32.6&72.6&83.5&188.7		&572.4\\
			\modelshort($\txtglobal_{sent}$+$\txtglobal_{obj}$) & 42.9&\textbf{77.2}&\textbf{85.6}&205.7		&30.1&69.0&79.8&178.9		&34.0&71.2&83.6&188.8		&573.4\\
			\modelshort($\txtglobal_{sent}$+$\txtglobal_{attr}$) &40.1&73.9&83.3&197.3 	&\textbf{37.4}&\textbf{72.0}&\textbf{81.9}&\textbf{191.3}	&30.5&70.0&81.9&182.4		&571.0\\
			\modelshort($\txtglobal_{sent}$+$\txtglobal_{rel}$)&\textbf{45.4}&77.1&85.5&\textbf{208.0}		&29.2&68.1&78.5&175.8  		&\textbf{42.8}&\textbf{77.5}&\textbf{85.6}&\textbf{205.9}		&\textbf{589.7}\\
			\hline				
		\end{tabular}
	\end{center}
	\caption{\label{tab:adv_attack}Results on image-to-sentence retrieval task with text-domain adversarial attacks. For each caption, we generate 5 adversarial fake captions which do not match the images. Thus, the models need to retrieve 5 positive captions from 30,000 candidate captions.
	}
	} 
	\vspace{-0.15in}
\end{table*}

\subsectionnoid{Global image-level alignment.}
For relational triples $\txtglobal_{rel}$, semantic components aggregations $\txtglobal_{comp}$ and sentences $\txtglobal_{sent}$, their semantics usually cover multiple objects. Thus, we align them with the full image embedding $\imgglobal$ via bidirectional ranking losses as Eq.~(\ref{eq:vse})\footnote{Only textual negative samples are used for $\ell_{rel}$.}. The alignment loss is denoted as $\ell_{rel}, \ell_{comp}$ and $\ell_{sent}$, respectively.

We want to highlight that, during training, we separately align the two type of semantic representations of the caption, \ie, $\txtglobal_{sent}$ and $\txtglobal_{comp}$, with the image. This differs from the inference-time computation of the caption. Recall that $\alpha$ can be viewed as a factor that balances the training objective and the enforcement of semantic coverage. This allows us to flexibly adjust $\alpha$ during inference.

\section{Experiments}
We evaluate our model on the MS-COCO \cite{mscoco} dataset. It contains 82,783 training images with each image annotated by 5 captions. We use the common 1K validation and test split from \cite{deepvs}. 

We first validate the effectiveness of enforcing the semantic coverage of caption embeddings by comparing models on cross-modal retrieval tasks with adversarial examples. We then propose a unified text-to-image retrieval task to support the contrastive learning on various semantic components. We end this section with an application of using visual cues to facilitate the semantic parsing of novel sentences. We include two baselines: VSE++\cite{vsepp} and VSE-C\cite{vsec} for comparison.

\subsection{Retrieval under text-domain adversarial attack}

Recent works \cite{vsec,foilit} have raised their concerns on the robustness of the learned visual-semantic embeddings. They show that existing models are vulnerable to text-domain adversarial attacks (\ie, using adversarial captions) and can be easily fooled. This is closely related to the bias in small datasets over a large, compositional semantic space \cite{vsec}. 
To prove the robustness of the learned unifed VSE, we further conduct experiments on the image-to-sentence retrieval task with text-domain adversarial attacks. Following \cite{vsec}, we first design several types of adversarial captions by adding perturbations to existing captions. 

\begin{enumerate}
    \item  \textbf{Object attack}: Randomly replace / append by an irrelevant one in the original caption. 
    \vspace{-0.05in}
    \item  \textbf{Attribute attack}: Randomly replace / add an irrelevant attribute modifier for one object in the original caption. 
    \vspace{-0.05in}
    \item \textbf{Relational attack}: 1) Randomly replace the subject/relation/object word by an irrelevant one. 2) Randomly select an entity as a subject/object and add an irrelevant relational word and object/subject. 
\end{enumerate}

The results are shown in Table~\ref{tab:adv_attack} where different columns represent different types of attacks. 
VSE++ performs worst as it is only optimized for the retrieval performance on the dataset. Its sentence encoder is insensitive to a small perturbation in the text.
VSE-C explicitly generates the adversarial captions based on human-designed rules as hard negative examples during training, which makes it relatively robust to those adversarial attacks. 
\model shows strong robustness across all types of adversarial attacks and outperforms all baselines.

The ability of \model to defend adversarial texts comes almost for free: we present {\it zero} adversarial captions during training. \model builds fine-grained semantic alignments via the contrastive learning of semantic components. It use the explicit aggregation of the components $\txtglobal_{comp}$ to alleviate the dataset biases.

\subsectionnoid{Ablation study: semantic components.}
We now delve into the effectiveness of different semantic components by choosing different combinations of components for the caption embedding.
Shown in Table~\ref{tab:adv_attack}, we use different subsets of the semantic components to form the bag-of-component embeddings $\txtglobal_{comp}$. For example, in \modelshortnopostspace$_{obj}$, only object nouns are selected and aggregated as $\txtglobal_{comp}$.

The results demonstrate the effectiveness of the enforcement of semantic coverage: even if the semantic components have got fine-grained alignment with visual concepts, directly using $\txtglobal_{sent}$ as the caption encoding still degenerates the robustness against adversarial examples.
Consistent with the intuition, enforcing of coverage of a certain type of components (\eg, objects) helps the model to defend the adversarial attacks of the same type (\eg, defending adversarial attacks of nouns). Combining all components leads to the best performance.

\begin{figure}[t]
	\centering
	\subfigure[Normal cross-modal retrieval (5,000 captions)]
	{\label{fig:alpha_normal}\includegraphics[width=38mm]{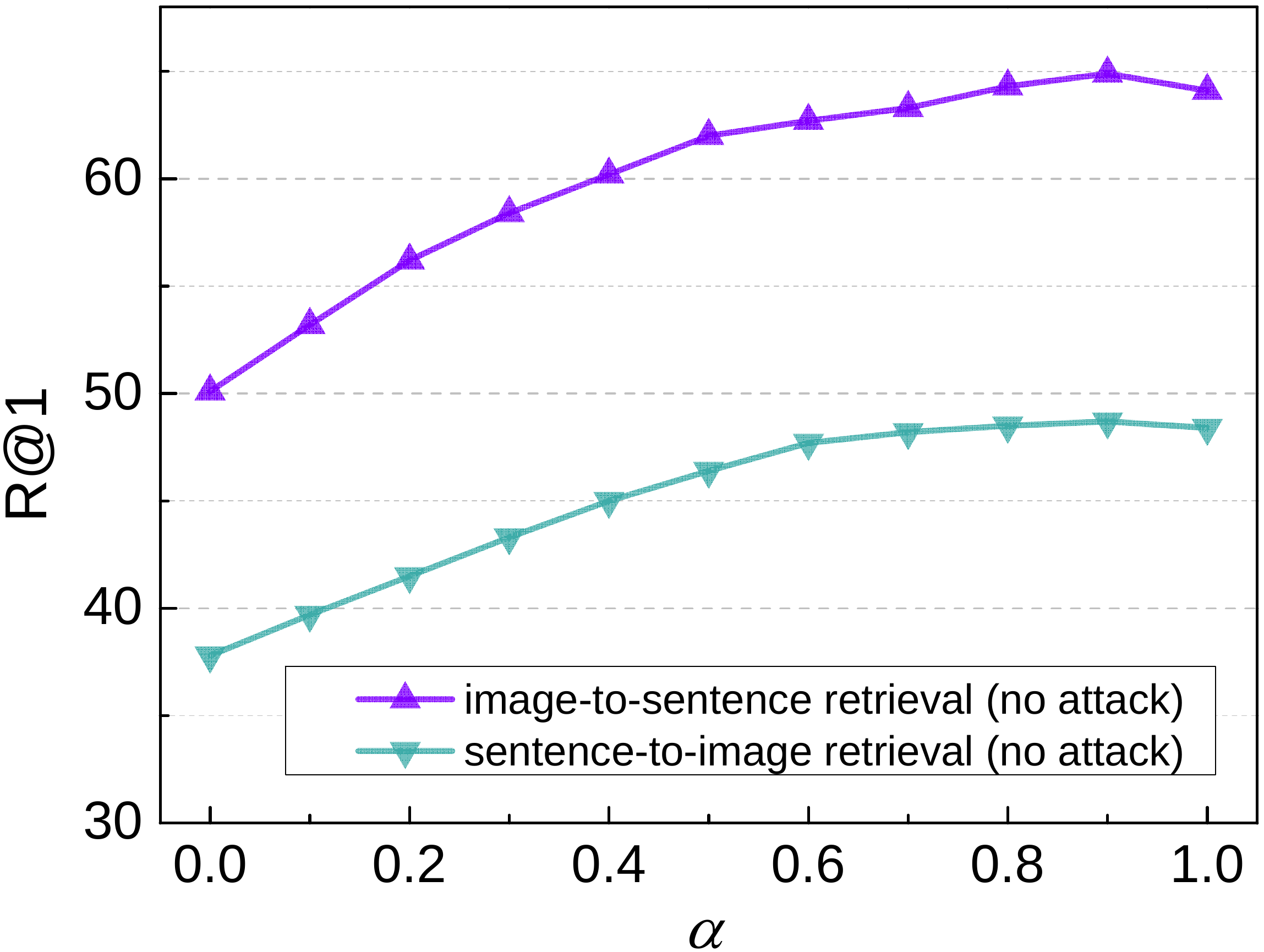}}
	\subfigure[Adversarial attacked image-to-sentence retrieval (30,000 captions)]
	{\label{fig:alpha_attack}\includegraphics[width=38mm]{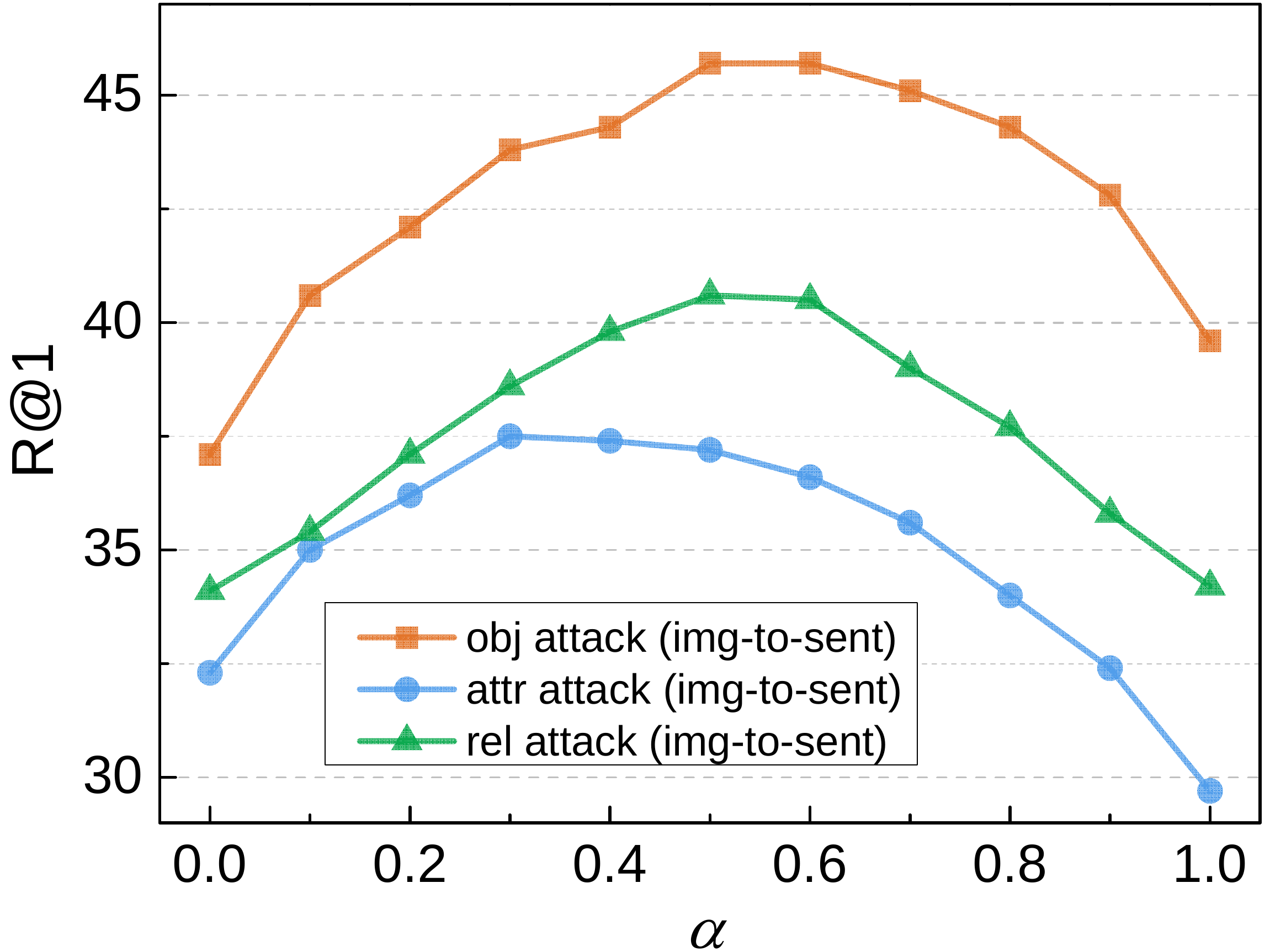}}\\
	\caption{\label{fig:alpha}The performance of \modelshort on cross-modal retrieval tasks with different combination weight $\alpha$. Our model can effective defending adversarial attacks, with no sacrifice for the performance on other tasks by choosing a reasonable $\alpha$ (thus we set $\alpha=0.75$ in all other experiments).
    }
\end{figure}
\subsectionnoid{Choice of the combination factor: $\alpha$.}
We study the choice of $\alpha$ by conducting experiments on both normal retrieval tasks and the adversarial one. Fig~\ref{fig:alpha} shows the R@1 performance under the normal/adversarial retrieval scenario \wrt different choices of $\alpha$.
We observe that the $\txtglobal_{comp}$ term contributes little on the normal cross-modal retrieval tasks but largely on tasks with adversarial attacks. 
Recall that $\alpha$ can be viewed as a factor that balances the training objective and the enforcement of semantic coverage.
By choosing $\alpha$ from a reasonable region (\eg, from 0.6 to 0.8), our model can effective defend adversarial attacks, with no sacrifice for the overall performance.

\begin{figure*}
	\centering
	\includegraphics[width=160mm]{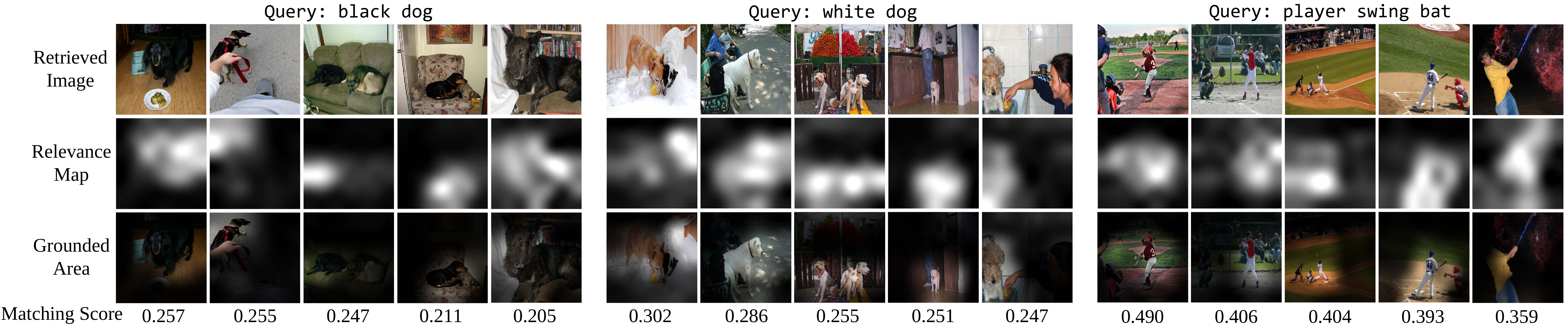}
	\caption{The relevance maps and grounded areas obtained from the retrieved images w.r.t. three queries. The temperature of the {\tt softmax} for visualizing the relevance map is $\tau=0.1$. Pixels in white indicates a higher matching score. Note that the third image of the query ``black dog'' contains two dogs, while our model successfully locates the black one (on the left). It also succeeded in finding the white dog in the first image of ``white dog''. Moreover, for the query ``player swing bat'', although there are many players in the image, our model only attend to the man swinging the bat.}
	\label{fig:local_retrieval}
	\vspace{-0.05in}
\end{figure*}

\begin{figure*}
	\centering
	\includegraphics[width=160mm]{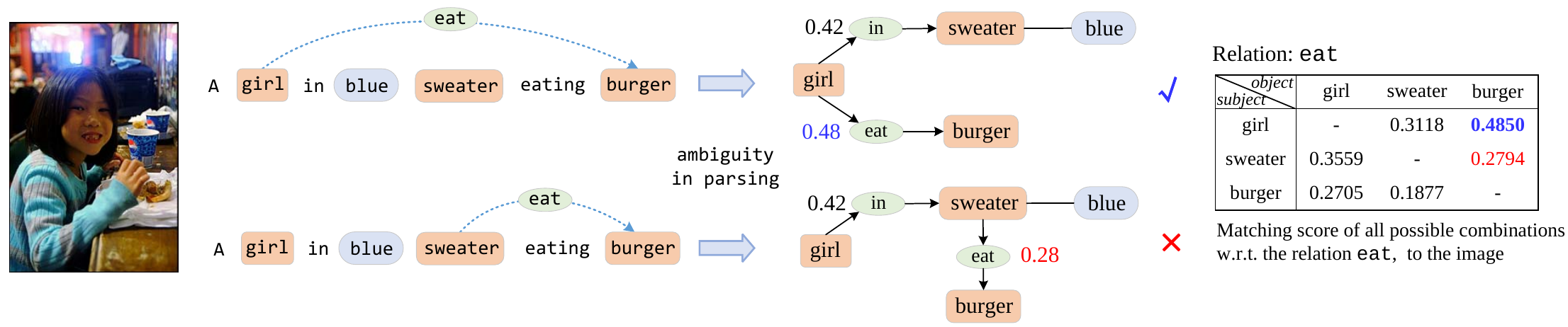}
	\caption{Example showing our model can leverage image to assist semantic parsing when there is ambiguity in the sentence. 
	We can infer that the matching score of ``girl eat burger'' is much higher than ``sweater eat burger'', which can help to eliminate the ambiguity. Note that the other components in the scene graph are also correctly inferred by our model.}
	\label{fig:syntactic_parsing}
	\vspace{-0.1in}
\end{figure*}

\subsection{Unified Text-to-Image Retrieval}
\begin{table}
	
	\begin{center}\scriptsize
		\begin{tabular}{c|c|c|c|c||c}
			\hline
			Task & obj & attr & rel  & obj (det) & sum \\
			\hline
			VSE++ & 29.95 & 26.64 & 27.54 & 50.57 & 134.70 \\
			VSE-C & 27.48 & 28.76 & 26.55  & 46.20  & 128.99\\
            \modelshortnopostspace$_{all}$ & \textbf{39.49} & \textbf{33.43} & \textbf{39.13} & \textbf{58.37}& \textbf{170.42}\\
            \hline
			\modelshortnopostspace$_{obj}$ & \textbf{39.71} & 33.37 & 34.38  & 56.84 & 164.3\\
			\modelshortnopostspace$_{attr}$ & 31.31 & \textbf{37.51} & 34.73 & 52.26 & 155.81\\
			\modelshortnopostspace$_{rel}$ & 37.55 & 32.7 & \textbf{39.57} & \textbf{59.12}  & \textbf{168.94} \\
			\hline
		\end{tabular}
	\end{center}
	%
	\caption{\label{tab:local_retrieval} The mAP performance on the unified text-to-image retrieval task.
	Please refer to the text for details.
	}
\end{table}

We extend the word-to-scene retrieval used by \cite{vsec} into a general {\it unified text-to-image retrieval} task. In this task, models receive queries of different semantic levels, including single words (\eg, ``Clock.''), noun phrases (\eg, ``White clock.''), relational phrases (\eg, ``Clocks on wall'') and full sentences. For all baselines, the texts of different types as treated as full sentences. The result is presented in Table~\ref{tab:local_retrieval}.

We generate positive image-text pairs by randomly choosing an image and a semantic component from 5 matched captions with the chosen image.
It is worth mention that the semantic components extracted from captions may not cover all visual concepts in the corresponding image, which makes the annotation noisy. To address this, we also leverage the MS-COCO detection annotations to facilitate the evaluation (see \emph{obj(det)} column). We treat the labels for detection bounding boxes as the annotation of objects in the scene.

\subsectionnoid{Ablation study: contrastive learning of components.}
We evaluate the effectiveness of using contrastive samples for different semantic components. Shown in Table~\ref{tab:local_retrieval}, \modelshortnopostspace$_{obj}$ denotes the model trained with only contrastive samples of noun components. The same notation applies to other models. The \modelshort trained with a certain type of contrastive examples (\eg, \modelshortnopostspace$_{obj}$ with contrastive nouns) consistently improves the retrieval performance of the same type of queries (\eg, retrieving images from a single noun). \modelshort trained with all kinds of contrastive samples performs best in overall and shows a significant gap \wrt other baselines.

\subsectionnoid{Visualization of the semantic alignment.}
We visualize the semantic-relevance map on an image \wrt a given query $\txtglobal_q$ for a qualitative evaluation of the alignment performance of various semantic components.
The map $\mathbf{M}_i$ is computed as the similarity between each image region $\imgglobal_i$ and $\txtglobal_q$, in a similar way as Eq. (\ref{eq:relevance_map}). Shown as Fig.~\ref{fig:local_retrieval}, this visualization helps to verify that our model successfully aligns different semantic components with the corresponding image regions.

\subsection{Semantic Parsing with Visual Cues}
As a side application, we show how the learned unified VSE space can provide the visual cues to help the semantic parsing of sentences. Fig.~\ref{fig:syntactic_parsing} shows the general idea. When parsing a sentence, ambiguity may occur, \eg, the subject of the relational word \verb`eat` may be \verb`sweater` or \verb`burger`. It is not easy for a textual parser to decide which one is correct because of the innate syntactic ambiguity. However, we can use the image which is depicted by this sentence to assist the parsing by. This is related to previous works on using image segmentation models to facilitate the sentence parsing \cite{visual_cue}.

This motivates us to design two tasks, 1) recovering the dependency between attributes and entities, and 2) recovering the relational triples.
In detail, we first extract the entities, attributes and relational words from the raw sentence without knowing their dependencies. For each possible combination of certain semantic component, our model computes its embedding in the unified joint space. \Eg, in Fig.~\ref{fig:syntactic_parsing}, there are in total $3\times (3 - 1) = 6$ possible dependencies for \verb`eat`. We choose the combination with the highest matching score with the image to decide the subject/object dependencies of the relation \verb`eat`.
We use parsed semantic components as the ground-truth and report the accuracy, defined as the fraction of the number of correct dependency resolution and the total number of attributes/relations.

Table~\ref{tab:syntactic_parsing} reports the results on assisting semantic parsing with visual cues, compared with other baselines. Fig.~\ref{fig:syntactic_parsing} shows a real case in which we successfully resolve the textual ambiguity.

\begin{table}
\vspace{-0.1in}
	\small
	\begin{center}
		\begin{tabular}{c|c|c}
			\hline
			Task & attributed object & relational phrase\\
			\hline
			Random & 37.41 & 31.90\\
			VSE++ & 41.12 & 43.31\\
			VSE-C & 43.44 & 41.08\\
			\modelshort & \textbf{64.82} & \textbf{62.69} \\
			\hline
		\end{tabular}
	\end{center}
	\caption{\label{tab:syntactic_parsing} The accuracy of different models on recovering word dependencies with visual cues. In the ``Random'' baseline, we randomly assign the word dependencies.}
	\vspace{-0.15in}
\end{table}

\vspace{-0.1in}
\section{Conclusion}
We present a {\it unified visual-semantic embedding} approach that learns a joint representation space of vision and language in a factorized manner: Different levels of textual semantic components such as objects and relations get aligned with regions of images.
A contrastive learning approach for semantic components is proposed for the efficient learning of the fine-grained alignment.
We also introduce the enforcement of semantic coverage: each caption embedding should have a coverage of all semantic components in the sentence. \model shows superiority on multiple cross-modal retrieval tasks and can effectively defend text-domain adversarial attacks.
We hope the proposed approach can empower machines that learn vision and language jointly, efficiently and robustly.

\bibliography{naaclhlt2019}
\bibliographystyle{acl_natbib}

\end{document}